\documentclass{article}
\usepackage{spconf}
\usepackage{textcomp}
\usepackage{graphicx}
\usepackage{subcaption}
\usepackage[sorting=none]{biblatex}
\usepackage[absolute]{textpos}
\usepackage[table]{xcolor}

\title{Inter and Intra-Annual Spatio-Temporal Variability of \\Habitat Suitability for Asian Elephants in India: \\A Random Forest Model-based Analysis}

\twoauthors{P. Anjali}
{Dept. of Computational and Data Sciences\\
Indian Institute of Science\\
Bangalore, India 560 012\\
anjalip1@iisc.ac.in}
           {Deepak N. Subramani}
{Dept. of Computational and Data Sciences\\
Indian Institute of Science\\
Bangalore, India 560 012\\
deepakns@iisc.ac.in}
            
\begin{document}

\begin{textblock}{13}(1.6,1)
\noindent Submitted for possible publication in the IEEE International India Geoscience and Remote Sensing Symposium 2021 (InGARSS 2021), scheduled for December 6 – 10, 2021 in Ahmedabad, Gujarat, India.
\end{textblock}

\maketitle
\ninept

\begin{abstract}
We develop a Random Forest model to estimate the species distribution of Asian elephants in India and study the inter and intra-annual spatiotemporal variability of habitats suitable for them. Climatic, topographic variables and satellite-derived Land Use/Land Cover (LULC), Net Primary Productivity (NPP), Leaf Area Index (LAI), and Normalized Difference Vegetation Index (NDVI) are used as predictors, and the species sighting data of Asian elephants from Global Biodiversity Information Reserve is used to develop the Random Forest model. A careful hyper-parameter tuning and training-validation-testing cycle are completed to identify the significant predictors and develop a final model that gives precision and recall of 0.78 and 0.77. The model is applied to estimate the spatial and temporal variability of suitable habitats. We observe that seasonal reduction in the suitable habitat may explain the migration patterns of Asian elephants and the increasing human-elephant conflict. Further, the total available suitable habitat area is observed to have reduced, which exacerbates the problem. This machine learning model is intended to serve as an input to the Agent-Based Model that we are building as part of our Artificial Intelligence-driven decision support tool to reduce human-wildlife conflict.

\end{abstract}

\section{Introduction}
Asian elephants (\textit{Elephas maximus}) once ranged over a large part of the Asian continent but are now restricted to highly fragmented zones in most parts while being extinct in the others. The declining population of this endangered species requires strategic conservation measures to improve both the quantity and quality of the habitat, which has degraded severely owing to anthropogenic pressures. Under such a scenario, habitat suitability studies play a crucial role in planning and conservation actions and tackling issues such as Human-Wildlife conflict. As per the environment ministry data, a total of 655 elephants were killed throughout India from 2009 to 2016 due to conflict with humans \cite{mayank}. The main reasons for the deaths were reported to be electrocution, train accident, poaching, and poisoning. The years 2010-11 and 2012-13 topped in elephant deaths when 106 and 105 deaths, respectively, were reported.

The widely used MaxEnt method \cite{maxent} uses presence-only data to estimate species distribution, but fails terribly owing to the absence of sufficient data during all years and months which can adequately represent the environmental covariates to model the spatio-temporal change. Also, in the prevailing situation of drastic climate changes and shrinking forest land, studies about seasonal habitat changes is the need of the hour for understanding movement and migration patterns of the species. The advent of hyper-spectral sensors provides immense possibilities in the field of geosciences to monitor the evolution of ecosystems over time. This sensing technique can be utilized to arrive at a species habitat suitability model which can then be used for planning and conservation measures.

The present paper models the inter and intra-annual spatio-temporal variability in the habitat suitability of \textit{E. maximus} using a machine learning approach. The newly developed model is used to study habitat degradation during the period 2001-2016. This analysis is the first step towards developing a comprehensive Artificial Intelligence (AI) based model for suggesting policies for Human-Elephant conflict in India.

\section{Data and Methods} \label{sec:data}

\textbf{Data}. The target variable for the present study is the species presence data of \textit{E. maximus}. This data set is obtained from the Global Biodiversity Information Reserve \cite{GBIF}. A total of 231 observations of \textit{E. maximus} were available from 2000 to 2016. There is an uncertainty associated with each observation location. Therefore two more pseudo-presence locations were sampled from the locations within the uncertainty range. We use predictor variables from three categories: climatic, topographic, and vegetation-related. Table \ref{tab:data} provides a summary of the variables and their sources.

\textbf{Methods}. We use the Random Forest (RF) Classifier for modelling habitat suitability \cite{Breiman}. 
Random Forest is a versatile model with the ability to capture complex relationships between environmental variables and species presence \cite{Valavi}. Further, Random Forest is an ensemble learning scheme with randomness introduced in the feature selection process for each tree and in the choice of features for each split. As such, it is expected to outperform other machine learning algorithms. The classical species distribution model MaxEnt \cite{maxent} suffers from an issue due to the limited number of observation data, making it difficult to model the time periods for which data was unavailable. On the other hand, the Random Forest model enables us to develop the relationship between the environment variables and species presence in a general setting, which can predict the habitat suitability even for time periods when no species presence observation data is available.

\begin{table*}[t]
\centering
\caption{Summary of the predictor variables used in the Random Forest habitat suitability model} \label{tab:data}
\begin{tabular}{ |p{1.8cm}|p{4.8cm}|p{5cm}|p{2cm}| p{1.5cm}| }
\hline 
\cellcolor{gray!25}\textbf{Category} &  \cellcolor{gray!25}\textbf{Variable} &  \cellcolor{gray!25}\textbf{Source} &  \cellcolor{gray!25}\textbf{Unit} &  \cellcolor{gray!25}\textbf{Spatial resolution} \\
\hline
\hline
Climatic   & Monthly precipitation & https://www.worldclim.org & mm & 2.5 minute\\
  \hline
 &   Monthly minimum temperature & https://www.worldclim.org & \textdegree{}C  & 2.5 minute\\
  \hline
  &   Monthly maximum temperature & https://www.worldclim.org & \textdegree{}C & 2.5 minute\\
 \hline
 \hline
  Topographic  & Elevation above sea level & https://www.worldclim.org & m & 30 arc-seconds\\
  \hline
 &   Distance to rivers and water-bodies & Derived using QGIS with data downloaded from https://www.openstreetmap.org  & m &\\
  \hline
  &   Distance to roads &  Derived using QGIS with data downloaded from https://www.openstreetmap.org & m &\\
 \hline
   &   Land Use Land Cover (LULC) & https://bhuvan.nrsc.gov.in & categorical & 30m \\
 \hline
  \hline
  Vegetation related & Net Primary Productivity (NPP) & https://neo.sci.gsfc.nasa.gov & gC/m²/day & 0.1 degrees\\
  \hline
 &   Leaf Area Index (LAI) & https://neo.sci.gsfc.nasa.gov & m²/m² & 0.1 degrees\\
  \hline
  &   Normalized Difference Vegetation Index (NDVI) &  https://neo.sci.gsfc.nasa.gov & Dimensionless & 0.1 degrees\\
 \hline
\end{tabular}
\end{table*}

\section{Random Forest Model Development}
We first prepare a labeled data set to develop the Random Forest Classification model from the available data. The target is set as the presence, pseudo-presence and pseudo-absence data, for which the corresponding predictor variables are obtained. Pseudo-presence is required due to the uncertainty in the presence data (Sec.~\ref{sec:data}). Pseudo-absence data is generated since the random forest model requires both presence and absence data for training and the latter is unavailable \cite{RF}. Pseudo-absence points were sampled from regions outside the available IUCN range maps for \textit{E. maximus}, assuming the locations were unsuitable for occupancy \cite{Williams}. 70\% of the prepared labeled data set was used for training-validation and 30\% was used for testing or model evaluation.  

\begin{figure}[h!]
\includegraphics[width=\columnwidth]{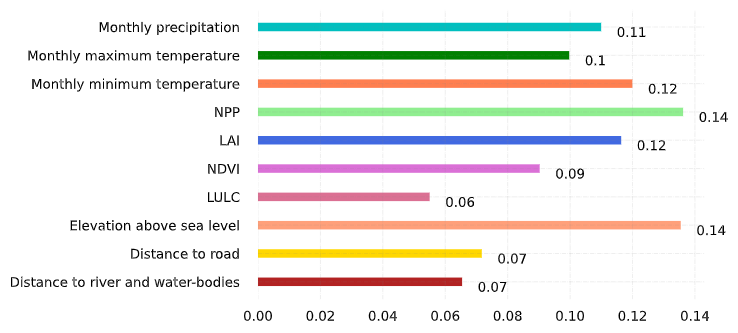}
\centering
\caption{Importance factor of the predictor variables used in the Random Forest model} \label{fig:importance}
\end{figure}

During model development, the predictor variables (from among the ten listed in Table \ref{tab:data}) with the least \textit{importance-factors}, viz. Land Use/Land Cover (LULC), distance to roads, and distance to rivers and water-bodies were identified and discarded in the final model. Fig.~\ref{fig:importance} shows a bar chart of the \textit{importance-factors} for the different predictor variables. The most significant predictor variables were Net Primary Productivity (NPP), Leaf Area Index (LAI) and elevation above sea level. NPP measures the flux of carbon dioxide between the plants and the environment, and is related to vegetation growth over time. LAI estimates the area covered with leaves over a unit land area, and thus measures the canopy thickness. The poor correlation of presence data with distance to roads and LULC could be due to the severe fragmentation of the habitat due to agricultural and developmental activities.

For regularizing the model, the hyper-parameters tuned were the number of estimators, number of features considered at every split, and maximum tree depth. The model was selected with 500 estimators, two features for splitting at every node, and a maximum tree depth of 22.

\begin{figure}[h!]
\includegraphics[width=\columnwidth]{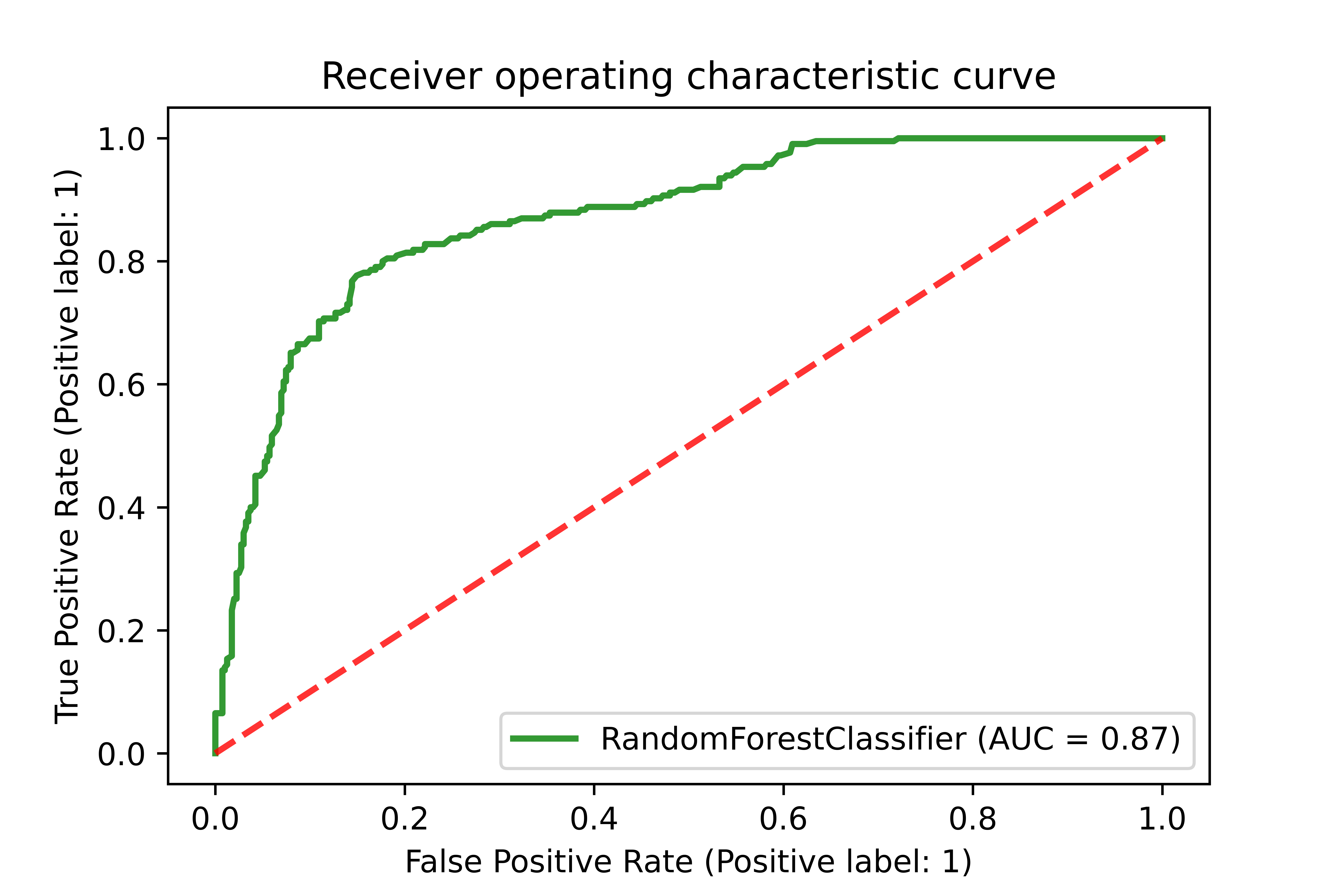}
\centering
\caption{ROC curve for the Random Forest model training} \label{fig:auc}
\end{figure}

The final model was selected as the model that had the best classification metrics, quantified through the area under the receiver operating characteristic curve (AUC; Fig.~\ref{fig:auc}) and precision-recall values. The testing precision for the best model is 0.77, recall is 0.76, and F1 score is 0.769, indicating that the model generalizes well and can reasonably estimate the suitable habitat.

\section{Spatio-temporal Analysis of Habitat Suitability}
\begin{figure}[b]
\includegraphics[width=8.5cm]{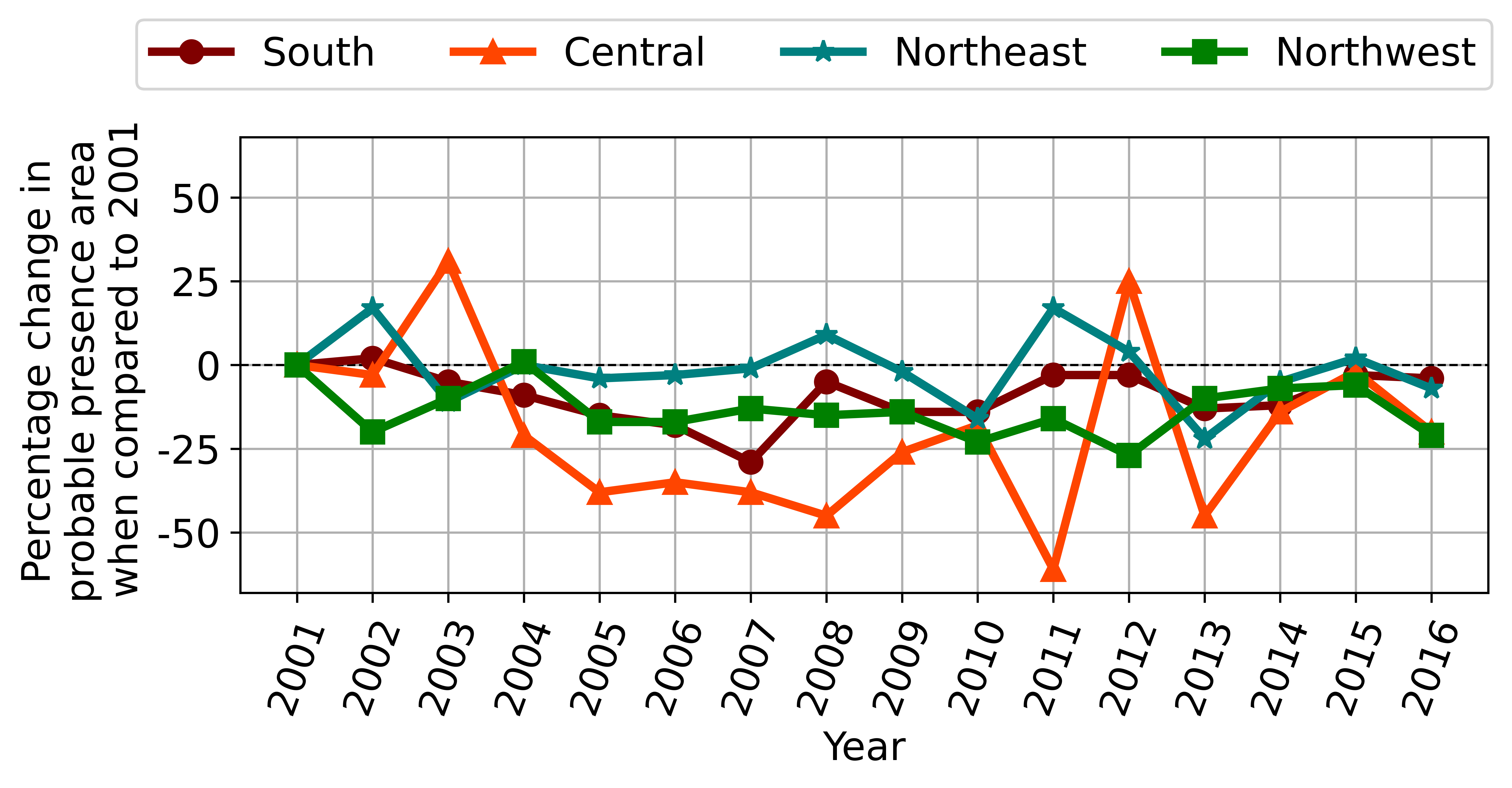}
\centering
\caption{The percentage change in the area classified as probable presence during 2001-2016 when compared with 2001 for the four regional zones in India} \label{fig:timeseries2}
\end{figure}
\begin{figure}[h!]
\includegraphics[width=8.5cm]{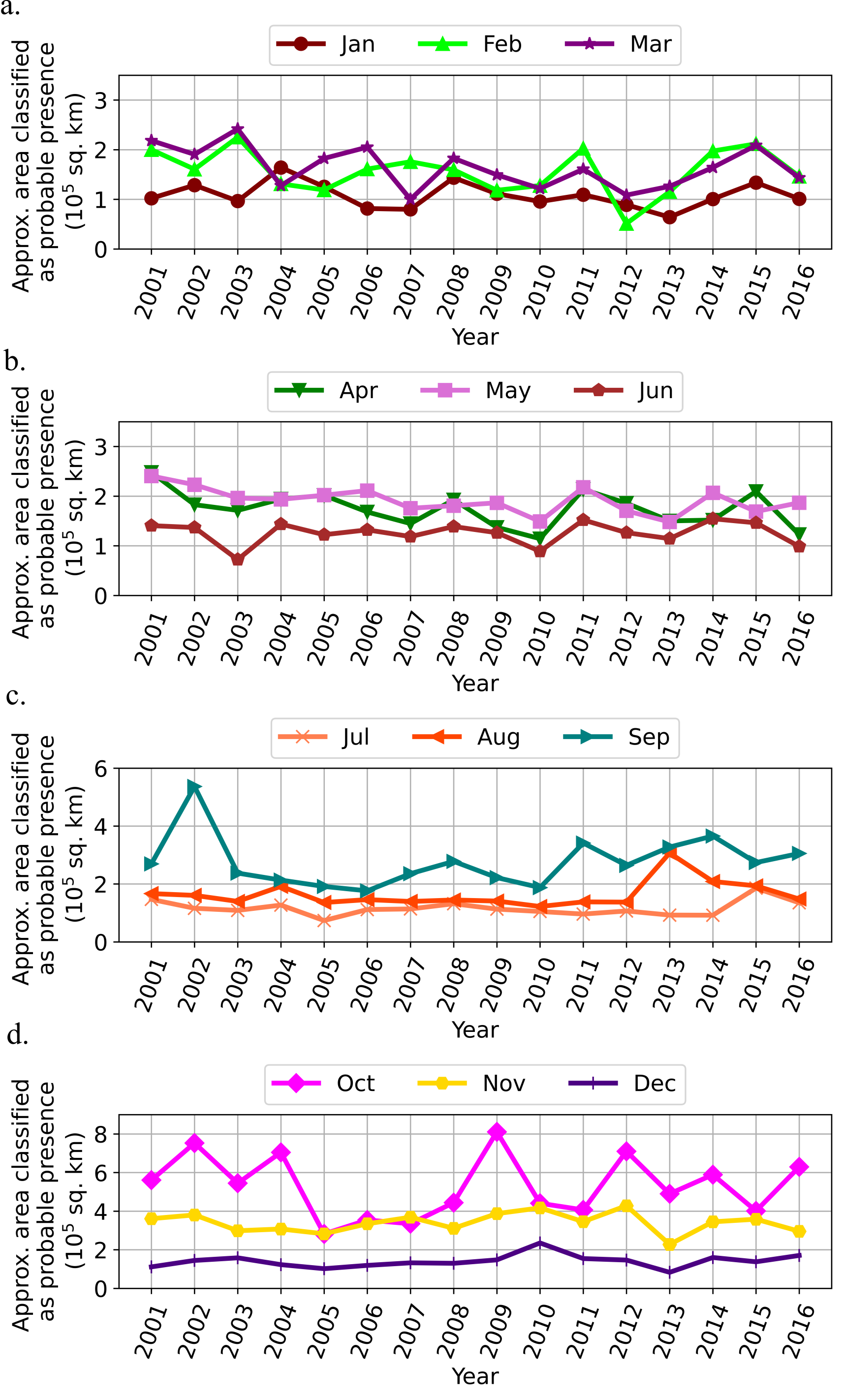}
\centering
\caption{Time series of the area (in sq. km) classified as a suitable habitat for Asian elephants by the Random Forest model during 2001-2016} \label{fig:timeseries1}
\end{figure}

\begin{figure*}
\includegraphics[width=17.8cm]{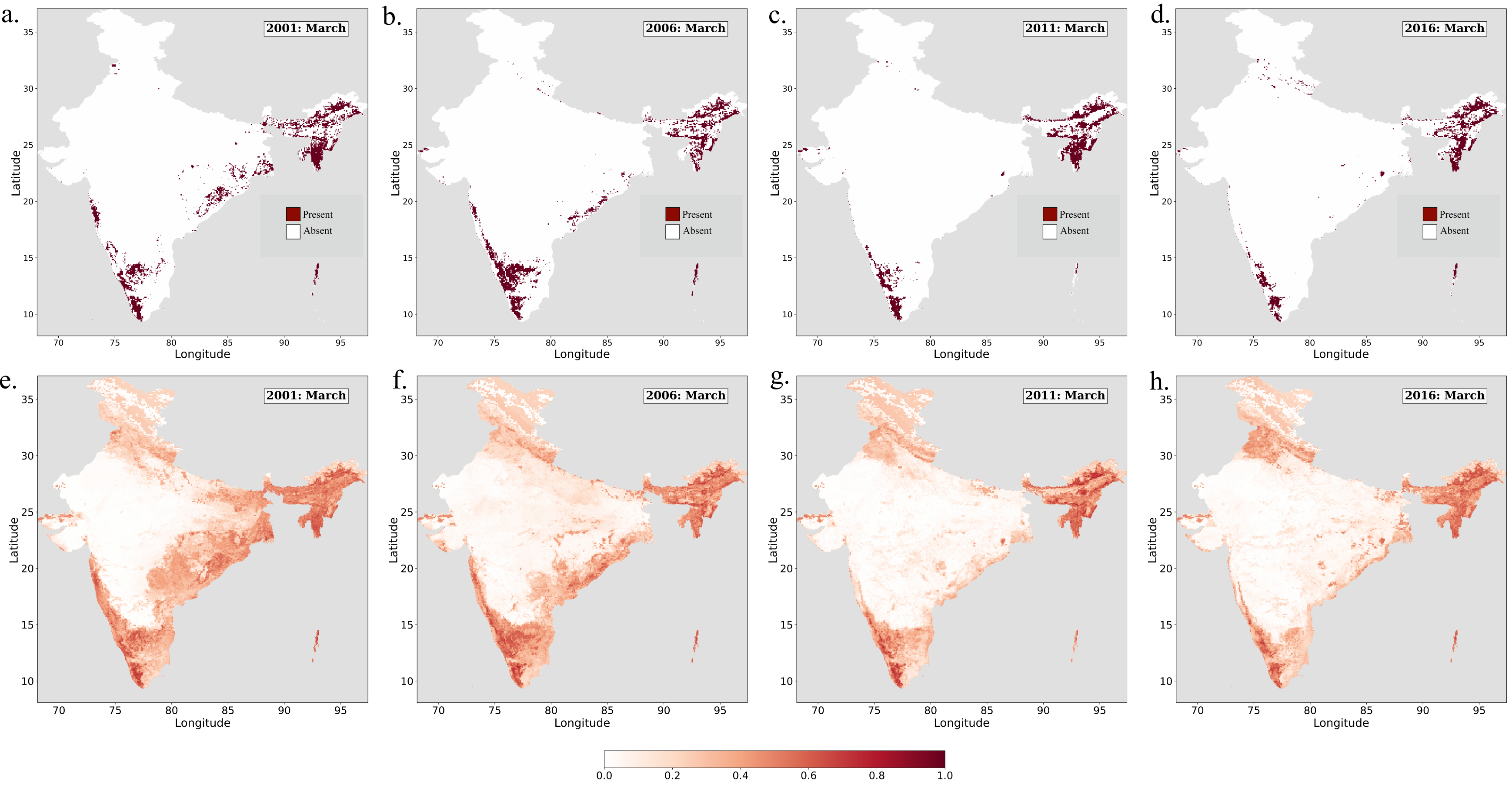}
\centering
\caption{Estimated change in the habitat suitability of \textit{E. maximus} for the month of March in the years 2001, 2006, 2011 and 2016 by the random forest model. Panels a, b, c, and d show the binary classification of estimated presence and absence. Panels e, f, g, and h show the probability distribution for classification of a pixel as presence, red indicating high habitat suitability and white indicating low habitat suitability} \label{fig:habitat}
\end{figure*}

Our Random Forest model was applied to estimate the suitable habitat range of \textit{E. maximus} in India from 2000 to 2016. The monthly mean predictor variables for each year were used as input and the corresponding output of the Random Forest model is used for the analysis. 

Spatially, \textit{E. maximus} occurs in four major areas in India: north-east, central, north-west and south India \cite{Sukumar}. Fig.~\ref{fig:timeseries2} shows the percentage change in the aforementioned four regions with respect to the base year of 2001. We observe a general decreasing trend with a few anomalous years in which the suitable habitat area increased. Crucially, the years 2011 and 2013, which saw the maximum decrease in habitat suitability, were the years that reported maximum elephant deaths \cite{mayank}. 

Fig.~\ref{fig:timeseries1} shows the inter and intra-annual variability of the total area of suitable habitat as estimated by the Random Forest model. Within a year, December to August season has the lowest area of suitable habitats. September to November months, following the monsoon season, have an increase in the suitable habitat areas. This trend is expected due to seasonal changes in climate and vegetation wherein the habitable area shrinks during the dry months while increased during the wet months. 

The spatial distribution of suitable habitat across the years in the month of March (Fig.~\ref{fig:habitat}) shows a downward trend. As can be seen, the suitable habitat range has shrunk significantly in the past few years. This shrinkage in suitable habitats could explain the inter-state movement and migration that has been observed in the recent years. Further, the spatial distribution shows severe fragmentation of habitat in the states of Kerala, Karnataka, and Tamil Nadu that could account for the migration of the elephants from these southern states to Goa and Maharashtra looking for better landscapes \cite{Williams}. A similar movement of elephants is also observed into the northern parts of Andhra Pradesh from Odisha \cite{Williams}, and can be explained by looking at the spatial pattern. Even though Western Ghats and Northeast exhibits relatively stable habitats, there is severe fragmentation of the habitat in central and north-western India, which are the regions that top the human elephant conflict problem in India. Such changes in the suitable habitat exacerbates the Human-Elephant conflict problem and must be carefully studied and policy changes implemented.

\section{Conclusion and Future Work}
A random forest model was developed to estimate the suitable habitat for \textit{E. maximus} in India by assembling a data set from remotely sensed products and species presence data. This model was developed as a sub-model for an agent-based modeling system for studying the human-wildlife conflict problem, emphasizing elephants in India. The shrinking habitats pose a severe threat to the existence of the species. Severe habitat fragmentation coupled with the ability of \textit{E. maximus} to traverse over long distances in search of food and shelter is expected to aggravate the already grim human-elephant conflict problem. With the population on the declining trend, it is crucial to develop strategic conservation actions to save this endangered species. Actions for preserving existing habitats and ensuring connectivity by maintaining the corridors between them should be taken up imminently and seriously. Our next step is to develop an AI-based agent-based modeling system to aid in the decision-making to solve human-wildlife conflict problems.

\printbibliography

\end{document}